\documentclass[9pt,conference]{IEEEtran}
\IEEEoverridecommandlockouts
\usepackage{cite}
\usepackage{amsmath,amssymb,amsfonts}
\usepackage{algorithmic}
\usepackage{graphicx}
\usepackage{textcomp}
\usepackage{xcolor}
\usepackage{booktabs}
\usepackage{multirow}
\def\BibTeX{{\rm B\kern-.05em{\sc i\kern-.025em b}\kern-.08em
    T\kern-.1667em\lower.7ex\hbox{E}\kern-.125emX}}
\begin{document}

\title{Multilevel Semantic-Aware Model for AI-Generated Video Quality Assessment\\
}

\author{
\IEEEauthorblockN{Jiaze Li\textsuperscript{\dag}\thanks{\IEEEauthorrefmark{1} Equal contribution, \IEEEauthorrefmark{4}Corresponding author}\IEEEauthorrefmark{1}, 
Haoran Xu\textsuperscript{\dag}\IEEEauthorrefmark{1}, 
Shiding Zhu\textsuperscript{\dag}\IEEEauthorrefmark{1}, 
Junwei He\IEEEauthorrefmark{2}, 
Haozhao Wang\IEEEauthorrefmark{3}\IEEEauthorrefmark{4}}
\IEEEauthorblockA{\textsuperscript{\dag}\textit{Zhejiang University, Hangzhou, China}}
\IEEEauthorblockA{\IEEEauthorrefmark{2}\textit{University of Chinese Academy of Sciences, Beijing, China}}
\IEEEauthorblockA{\IEEEauthorrefmark{3}\textit{School of Computer Science and Technology, Huazhong University of Science and Technology, Wuhan, China}}
}

\maketitle
\begin{abstract}
The rapid development of diffusion models has greatly advanced AI-generated videos in terms of length and consistency recently, yet assessing AI-generated videos still remains challenging. Previous approaches have often focused on User-Generated Content(UGC), but few have targeted AI-Generated Video Quality Assessment methods. In this work, we introduce MSA-VQA, a Multilevel Semantic-Aware Model for AI-Generated Video Quality Assessment, which leverages CLIP-based semantic supervision and cross-attention mechanisms. Our hierarchical framework analyzes video content at three levels: frame, segment, and video. We propose a Prompt Semantic Supervision Module using text encoder of CLIP to ensure semantic consistency between videos and conditional prompts. Additionally, we propose the Semantic Mutation-aware Module to capture subtle variations between frames. Extensive experiments demonstrate our method achieves state-of-the-art results.
\end{abstract}


\section{INTRODUCTION}
\label{sec:intro}
\setlength{\parskip}{0pt}
In the rapidly evolving digital era, the demand for sophisticated Video Quality Assessment (VQA) is becoming increasingly paramount, especially in the context of AI-Generated Content (AIGC). As AIGC videos grow in prevalence, there is a pressing demand for VQA methodologies capable of accurately assessing the perceptual quality of these videos, which often diverge significantly from traditional Professional Generated Content (PGC) and User Generated Content (UGC).

Traditionally, VQA techniques have been classified into three main categories: full-reference (FR), reduced-reference (RR), and no-reference (NR), depending on the availability of a reference video \cite{ma2021image}. Early FR-VQA methods relied on pixel-wise comparisons to determine quality metrics, while RR-VQA approaches used partial reference data for evaluation. NR-VQA has gained prominence, particularly because of its relevance in situations where a pristine reference video is unavailable \cite{min2020study, mittal2015completely, saad2014blind, xu2024shortform}. These methods often employ handcrafted features, such as discrete cosine transformation coefficients \cite{li2016spatiotemporal} and optical flow \cite{manasa2016optical}, to statistically represent video quality.

The rise of deep learning has fundamentally transformed the VQA field, with convolutional neural networks (CNNs) becoming the dominant tool for feature extraction \cite{he2016deep, ren2016faster, liu2022video}. Notable innovations such as V-CORNAIA \cite{xu2014no}, DeepBVQA \cite{ahn2018deep}, and RIRNet \cite{chen2020rirnet} highlight the effectiveness of CNNs in identifying intricate patterns that correlate with video quality. Additionally, models like SimpleVQA \cite{sun2022deep} integrate spatial attributes derived from CNNs with temporal features from action recognition networks, effectively navigating the spatio-temporal dynamics unique to video content.

Despite these advances, conventional VQA models are primarily tailored to PGC and UGC videos, whose characteristics differ significantly from AI-Generated videos. AIGC videos often display distinctive features, such as alignment with specific textual prompts and sudden variations in content or quality, posing new challenges for traditional VQA frameworks.

To address these challenges, we propose MSA-VQA, a Multilevel Integration Model designed for AI-Generated Video Quality Assessment. Our model builds on insights from Zoom-VQA \cite{zhao2023zoom} and SimpleVQA \cite{sun2022deep} with introducing Prompt Semantic Supervision Module, multilevel feature extraction Module and Semantic Mutation-aware Module to evaluate AI-Generated videos. In summary, the key contributions of our model architecture are as follows:
\begin{itemize}
    \item We assess videos at three distinct levels: frame, segment, video and design specialized loss functions for each level, capturing video information more comprehensively.
    \item To determine whether the generated videos align with the conditional prompts, we introduce the Prompt Semantic Supervision Module in each model branch. This module employs CLIP's text encoder to process the conditional prompt and integrates it as a feature during model training.
    \item We introduce the Semantic Mutation-aware Module that leverages CLIP's image encoder to encode each video frame and applies cross-attention to evaluate semantic changes between frames.
    \item Extensive experimental validation demonstrates that our model achieves state-of-the-art performance.
\end{itemize}

\section{PROPOSED METHOD}
\begin{figure*}
    \centering
    \includegraphics[width=1\linewidth]{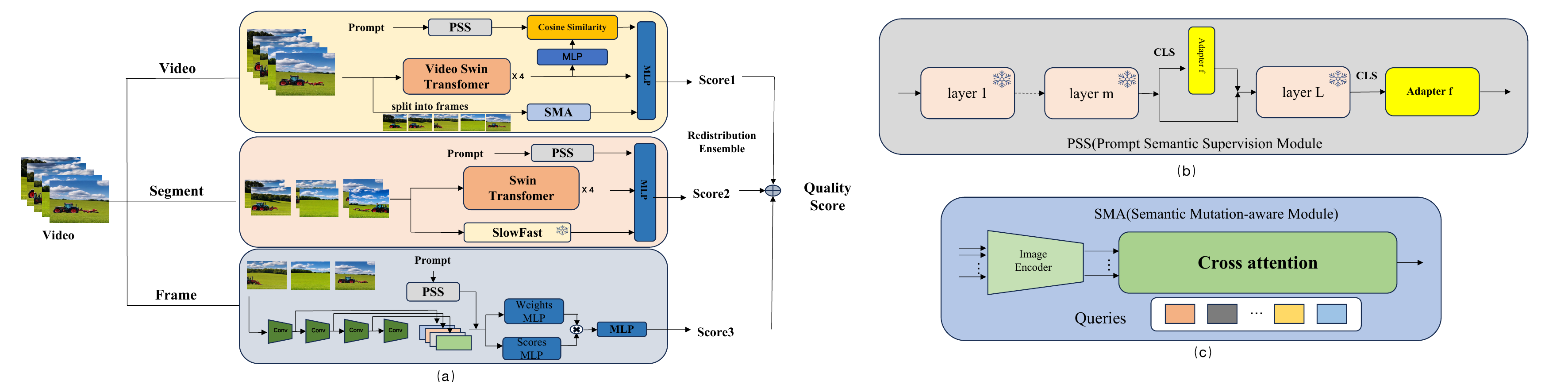}
    \caption{Illustration of the MSA-VQA framework. The framework includes three main components capturing features at the video, segment, and frame levels, as shown in (a). These components are trained separately for stability and ensembled during inference. A Prompt Semantic Supervision (PSS) module, based on the CLIP text encoder, ensures semantic alignment between the AI-Generated video and the prompt, as shown in (b). The Semantic Mutation-aware (SMA) Module models the semantic mutations between video frames, as indicated in (c).}
    \label{fig:enter-label}
\end{figure*}

Our proposed MSA-VQA framework is tailored for the quality assessment of videos within the context of AIGC. As depicted in Figure \ref{fig:enter-label}, the framework performs a multi-dimensional analysis of video quality across frame, segment, and video levels. To enhance the model's robustness and address the inherent variability of AIGC videos, data augmentation is systematically applied at both the frame and segment levels. The framework integrates the Prompt Semantic Supervision (PSS) module to assess the semantic alignment between videos and their prompts, as elaborated in Section \ref{subsec:PSS}. Additionally, the Semantic Mutation-aware (SMA) Module is introduced to detect mutation semantic transitions between frames, further discussed in Section \ref{subsec:SMA}. The integration of these multi-level features, along with the custom loss functions employed for model training, is thoroughly described in Section \ref{subsec:ensemble}.

\subsection{Data Augmentation}
\label{subsec:data_augmentation}
\subsubsection{Frame-Level}
\label{subsec:Frame_Level}
In the domain of video quality assessment (VQA), evaluation granularity is pivotal for precision. We employ frame-level data augmentation to enhance video quality analysis, treating each frame as a fundamental unit for detailed assessment. Our approach converts video-level quality scores to a frame-specific score by distributing the overall video quality scores across its frames, enriching each frame with the video's quality indicators. This method not only increases the volume of training data but also improves the model's sensitivity to intricate details.

We utilize a deep convolutional architecture to detect local features and textures critical for perceptual quality judgment, thereby enhancing the model's ability to identify complex video qualities through frame-level augmentation. 


\subsubsection{Segment-Level}
\label{subsec:Segment_Level}

In this research, we improved the performance of our SimpleVQA architecture by integrating the Swin Transformer \cite{liu2021swin}, pre-trained on the LSVQ dataset, as the foundation for our segment-level augmentation strategy, aimed at enhancing video analysis robustness and accuracy. Our methodology involves segmenting videos into units for focused analysis, allowing for precise enhancements tailored to the segment-specific features.

A key element of our strategy is the random initialization of segment start points to prevent model bias, thereby increasing its generalizability across diverse scenarios and enabling learning from varied temporal contexts. We further refine our approach through spatial and temporal data alignment techniques to ensure feature consistency and sequence synchronization across segments, which is essential for accurate event sequencing and the understanding of transitions.

\subsection{Prompt Semantic Supervision Module}
\label{subsec:PSS}
CLIP \cite{radford2021learningtransferablevisualmodels} is pretrained on a large-scale dataset of image-text pairs, aligning the feature spaces of images and texts during training and demonstrating robust performance on downstream tasks. In current mainstream text-based video generation models, prompts are encoded by CLIP's text encoder and then fed into the diffusion model as semantic information to guide video generation. We observed that for AI-Generated video quality assessment, the greater the semantic difference between a video's content and the conditional prompt used for its generation, the lower its quality score. Based on this observation, we utilize the encoding of prompts by CLIP's text encoder to semantically assess the consistency between the generated video and its corresponding prompt.



We incorporated adapters into CLIP's text encoder to improve the semantic relevance of the encoded information, as depicted in Figure \ref{fig:enter-label}(c). Specifically, adapters were introduced after the final two layers of CLIP's text encoder. A video contional prompt, $T_i$, is transformed into $[F, CLS]$, where $CLS$ denotes the class token and $F \in \mathbb{R}^{N \times C}$ represents the semantic features of the prompt. The adapters, represented  as $g(\cdot)$, project the class token into a quality-aware space. This process is summarized as:
\begin{align}
P_{c1} = g_1(\text{CLS})
\end{align}
\begin{align}
P_c = g_2(\text{Encoder}([P_{c1}, F]))
\end{align}
Here, $g_1(\cdot)$ and $g_2(\cdot)$ represent the adapters in the penultimate and final layers of the encoder, respectively, and $P_c$ is the final prompt mapped to a quality-aware space, providing an effective feature for assessing semantic consistency in AI-Generated videos.

\subsection{Semantic Mutation-aware Module}
\label{subsec:SMA}
\begin{figure}
    \centering
    \vspace{-0.5cm}
    \includegraphics[width=0.5\linewidth]{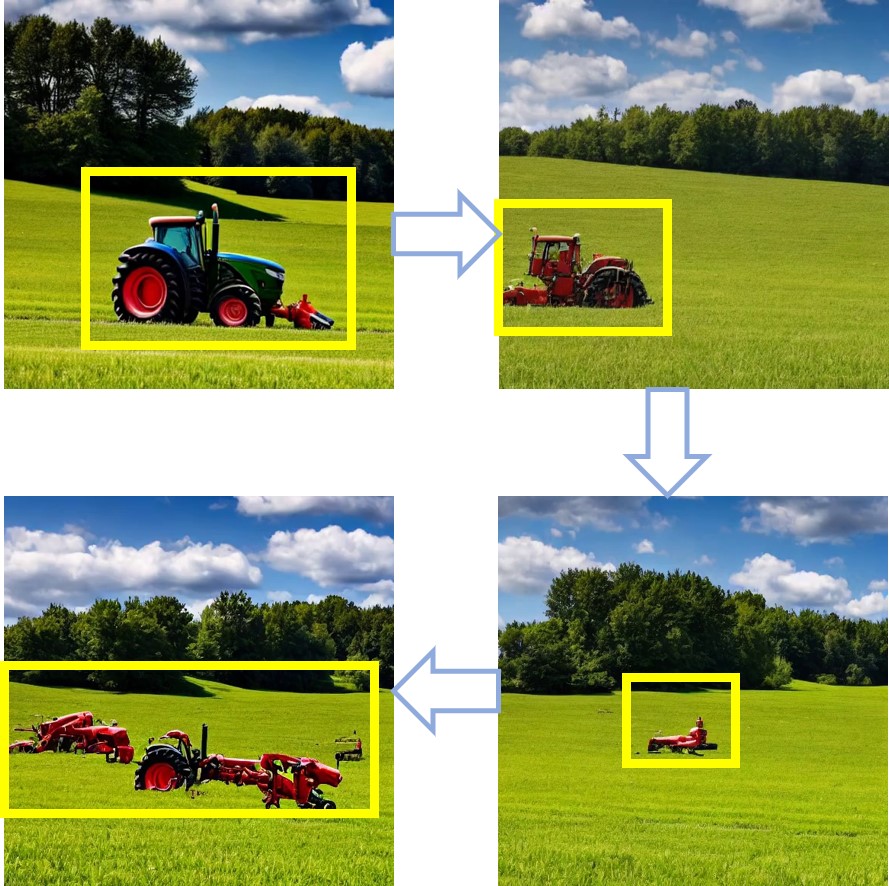}
    \caption{The four images above are from a video generated with the prompt: \textit{Time lapse of a field on which a tractor passes with a machine used to collect the cut grass and then make bales of hay, with the passage of white clouds on the blue sky}. The generated tractor (highlighted in yellow) shows significant instability and semantic mutations.}
    \label{fig:enter-label-2}
    \vspace{-0.5cm}
\end{figure}

As illustrated in Figure \ref{fig:enter-label-2}, we observe instances of semantic mutations between frames in AI-Generated videos, which significantly undermine the perceived realism and negatively impact the overall video quality. To effectively model these semantic mutations, we introduce the Semantic Mutation-aware Module (SMA). As shown in Figure \ref{fig:enter-label}(b), SMA leverages the image encoder from CLIP to extract the semantic information of video frames and employs learnable queries to detect semantic changes through cross-attention, enabling efficient learning. This design equips SMA with the essential ability to capture subtle semantic shifts between frames, thereby improving the accuracy of our video quality assessment approach.

For an input video $V_i$, we segment it into frames $[frame_1, ..., frame_n]$. Each frame is processed through CLIP’s encoder to obtain semantic features $[feature_1, ..., feature_n]$. We extract the CLS token from each feature, forming $IF = [CLS_1, ..., CLS_n]$. Since these features are not directly related to quality assessment, we apply adapters after the last two layers of CLIP to map them into a relevant feature space.

To model semantic mutations, SMA uses cross-attention to capture frame-to-frame variations, facilitating accurate video quality assessment. To address limited training data and reduce model complexity, we employ a learnable query to compress feature dimensions. Cross-attention uses $IF$  as key and value, and a trainable query $Q$ to learn semantic mutations, with the output represented as:
\begin{align}
F_{ca} &= CA(Q, IF, IF)
\end{align}
where $CA$ is cross-attention, $IF$ represents frame features, and $F_{ca}$ captures frame mutations. Both $F_{ca}$ and $Q$ share the same dimensions.
\subsection{Multilevel Model Ensemble Strategy}
\label{subsec:ensemble}
In this chapter, we explore Multi-level Model Ensemble (MME) strategies designed to integrate models trained at different data granularities: frame, segment, and video levels. Each branch operates independently during training and is paired with a specialized loss function to optimize performance for its data scope.

At the frame level, we use the smooth L1 loss, robust to outliers and preserving gradient information:
\begin{align}
L_{\text{smooth}} = 
\begin{cases} 
0.5 (\hat{y} - y)^2, & \text{if } |\hat{y} - y| < 1 \\ 
|\hat{y} - y| - 0.5, & \text{otherwise}
\end{cases} 
\end{align}
where $y$ and $\hat{y}$ are the ground truth and predicted values.

At the segment level, we combine mean absolute error (MAE) and rank loss to handle quantitative estimations and ordinal relationships:
\begin{align}
L_{\text{MAE}} = \frac{1}{N} \sum_{i=1}^{N} |y_i - \hat{y}_i|, 
\end{align}
\begin{align}
L_{\text{rank}} = \frac{1}{m^2} \sum_{i=1}^{m} \sum_{j=1}^{m} &\max(0, |y_i - y_j|  \notag \\ 
&- s(y_i, y_j)(\hat{y}_i - \hat{y}_j)) 
\end{align}
where $N$ is the number of segments, and $s(y_i, y_j) = 1$ if $y_i \geq y_j$, otherwise $-1$.

At the video level, we use Pearson Linear Correlation Coefficient (PLCC) loss:
\begin{align}
L_{\text{PLCC}} = 1 - \frac{\sum_{i=1}^{N}(y_i - \bar{y})(\hat{y}_i - \bar{\hat{y}})}{\sqrt{\sum_{i=1}^{N}(y_i - \bar{y})^2 \sum_{i=1}^{N}(\hat{y}_i - \bar{\hat{y}})^2}} 
\end{align}
where $\bar{y}$ and $\bar{\hat{y}}$ are the mean ground truth and predicted values.

After training, predictions are transformed using the sigmoid function:
\begin{align}
\sigma(z) = \frac{1}{1 + e^{-z}} 
\end{align}
and combined using a weighted sum:
\begin{align}
f_{ensemble}(x) = w_1 \cdot \sigma(z_1) + w_2 \cdot \sigma(z_2) + w_3 \cdot \sigma(z_3) 
\end{align}
where $w_1, w_2, w_3$ are optimized weights for ensemble predictions.

\section{EXPERIMENTS}
\subsection{Datasets and Evaluation Criteria}
In the field of AI-Generated video quality assessment, publicly available domain-specific datasets are limited. This paper utilizes T2VQA-DB datasets~\cite{Liu_2024_CVPR,kou2024subjectivealigneddatasetmetrictexttovideo}, dividing them into training and validation sets with a 9:1 ratio. The dataset contains 10,000 generated videos on 27 subjects from: Text2Video-Zero~\cite{text2video-zero}, AnimateDiff~\cite{guo2024animatediffanimatepersonalizedtexttoimage}, VideoFusion~\cite{luo2023videofusiondecomposeddiffusionmodels}, ModelScope~\cite{wang2023modelscopetexttovideotechnicalreport}, LVDM~\cite{he2022lvdm}, Show-1~\cite{zhang2023show1marryingpixellatent}. The video resolutionis unified to 512 x 512, and the video length is 4s. The LSVQ dataset~\cite{Sinno_2019}, the largest non-reference video quality assessment dataset, includes 39,000 videos with real-world distortions and 5.5 million human-annotated quality ratings. These annotations are crucial for evaluating and calibrating empirical models in video quality assessment (VQA). We use the LSVQ dataset to pre-train a segment-level Swin Transformer architecture, a state-of-the-art deep learning model for visual data processing. 


The model's performance is assessed using two key metrics: the Pearson Linear Correlation Coefficient (PLCC) and Spearman Rank Correlation Coefficient (SRCC). PLCC quantifies the linear relationship between predicted and actual values, with a value ranging from -1 (perfect negative correlation) to 1 (perfect positive correlation), and 0 indicating no linear relationship. The formula is given by:
\begin{align}
r = \frac{\sum_{i=1}^{N}(y_i - \bar{y})(\hat{y}_i - \bar{\hat{y}})}{\sqrt{\sum_{i=1}^{N}(y_i - \bar{y})^2 \sum_{i=1}^{N}(\hat{y}_i - \bar{\hat{y}})^2}}
\end{align}
where $\bar{y}$ and $\bar{\hat{y}}$ are the mean values of the ground truth and predictions, respectively.
The SRCC, on the other hand, measures the strength and direction of the monotonic relationship between two ranked variables. It is non-parametric and does not require the assumption of normality or a linear relationship. 
The formula for calculating $SRCC (\rho)$ is as follows:
\begin{align}
\rho = 1 - \frac{6 \sum_{i=1}^{n} d_i^2}{n(n^2 - 1)}
\end{align}
where $d_i$ is the difference in ranks between two variables for the i-th pair of observations, and $n$ is the total number of observations. $\rho$ is the SRCC value, also ranging from -1 to 1.
\subsection{Implementation details}
All experiments were conducted using PyTorch 2.0.0, with training accelerated by one NVIDIA A100 GPU. For the initial configuration of the Semantic Mutation-aware Module and Prompt Semantic Supervision Module, we used the ViT-B/32 version of CLIP. The architecture consists of three distinct branches, each following separate training protocols.The backbone and configurations of different branches are shown in Table \ref{tab:method}.

\begin{table}[h!]
    \centering
    \caption{Architectures and Configurations of Different Branches.}
    \label{tab:method}
    \resizebox{\linewidth}{!}{ 
    \begin{tabular}{c c c c c c}
    \toprule 
     Branch & Backbone & Optimizer & Input Res & LR & Batch Size \\
    \midrule 
    Video & Video Swin-trans. & AdamW & 336×336 & 1e-3 & 64\\
    Segment & Swin-trans. &  Adam & 448×448 & 1e-5 & 8\\
    Frame & ConvNext-tiny & AdamW & 320×320 & 2e-3 & 64\\
    \bottomrule 
    \end{tabular}
    }
\end{table}

\subsection{Performance Comparison with SOTA models}

\begin{table}[h!]
    \centering
    \caption{Comparison with SOTA methods. \textbf{Bold} fonts highlight the best performance.}
    \label{tab:competition}
    \begin{tabular}{c|c|cc|c}
    \toprule 
    Type&Method &SRCC$\uparrow$ &PLCC$\uparrow$& Avg Score$\uparrow$\\
    \midrule
    \multirow{5}{*}{zero-shot}&BLIP~\cite{li2022blipbootstrappinglanguageimagepretraining} & 0.165 & 0.186 & 0.175\\
    &ImageReward~\cite{xu2023imagereward} & 0.187 & 0.212 & 0.199\\
    &ViCLIP~\cite{wang2024internvidlargescalevideotextdataset} & 0.116 & 0.145 & 0.130\\
    &UMTScore~\cite{liu2023fetvbenchmarkfinegrainedevaluation} & 0.067 & 0.072 & 0.070\\
    &CLIPSim~\cite{radford2021learning} & 0.104 & 0.127 & 0.115\\
    \midrule 
    \multirow{2}{*}{handcrafted}&NIQE~\cite{6353522} & 0.549 & 0.625 & 0.587\\
    &FAVER~\cite{FAVER} & 0.648 & 0.692 & 0.672\\
    \midrule
    \multirow{9}{*}{finetuned}&SimpleVQA~\cite{sun2022deep} & 0.679 & 0.701 & 0.690\\
    &CLIP-IQA+~\cite{wang2022exploring} & 0.621 & 0.604 & 0.612\\
    &BVQA~\cite{li2022blindly} & 0.748 & 0.739 & 0.743\\
    &FAST-VQA~\cite{220wu2022fast} & 0.729 & 0.717 & 0.723\\
    &FasterVQA~\cite{10264158} & 0.745 & 0.722 & 0.734 \\
    &ZOOM-VQA~\cite{zhao2023zoom} & 0.725 & 0.756 & 0.740\\
    &DOVER~\cite{206wu2023exploring} & 0.672 & 0.691 & 0.682\\
    &Q-Align~\cite{wu2023qalign} & 0.759 & 0.748 & 0.753 \\
    &T2VQA~\cite{kou2024subjectivealigneddatasetmetrictexttovideo} &0.796&0.806&0.801\\
    \midrule
    finetuned &\textbf{Ours} & \textbf{0.810} & \textbf{0.825} & \textbf{0.818}\\
    \bottomrule 
    \end{tabular}
\end{table}

In this study, we rigorously compared our proposed method with current SOTA baselines. We compare our method with three different types of baseline: zero-shot, finetuned, and handcrafted. The results, as shown in Table \ref{tab:competition}, demonstrate a significant improvement across all evaluation metrics when our model is compared to top-performing existing models and achieve best performance on three different metrics.

Specifically, compared to SimpleVQA, which relies on temporal features for video quality assessment, our model showed a substantial increase in the avg score, from 0.690 to 0.818. Moreover, our method surpassed Zoom-VQA, which integrates both image and video features, by improving the SRCC from 0.725 to 0.810. Similarly, compared to DOVER~\cite{206wu2023exploring}, which decomposes quality assessment into technical and aesthetic aspects, our model significantly increased the PLCC from 0.691 to 0.825. Addtionally, we find that the performance of zero-shot and handcrafted method is often poor, possibly because the domain of the AI-Generated video is more variable, and therefore the challenge of handcrafted features or zero-shot approaches is greater and the ability to generalize on data sets in different domains is also worse. These results indicate that previous SOTA methods may struggle with the complexity of AI-Generated video quality assessment. In contrast, our MSA-VQA model incorporates multi-level features and evaluates both semantic consistency between video content and prompts, as well as the mutation of semantic changes across video frames. These components  guarantee the performance superiority of our method.

This comprehensive approach enables more accurate evaluation of AI-Generated video quality, addressing the limitations of previous models and setting a new benchmark for the field.

\subsection{Ablation Studies}
\begin{table}[h!]
    \centering
    \caption{Ablation study on different model ensemble strategies. \textbf{Bold} fonts highlight the best performance.}
    \label{tab:Ablation1}
    \begin{tabular}{c c c|cc|c}
    \toprule 
    Frame & Segment & Video & SRCC$\uparrow$ & PLCC$\uparrow$ & Avg Score$\uparrow$ \\
    \midrule 
    $\checkmark$ & $\checkmark$ & $\checkmark$ & \textbf{0.810} & \textbf{0.825} & \textbf{0.818} \\
    $\checkmark$ & $\times$ & $\checkmark$ & 0.787 & 0.809 & 0.798 \\
    $\checkmark$ & $\times$ & $\times$ & 0.725 & 0.756 & 0.741 \\
    $\times$ & $\checkmark$ & $\times$ & 0.746 & 0.774 & 0.760 \\
    $\times$ & $\times$ & $\checkmark$ & 0.764 & 0.790 & 0.777 \\
    \bottomrule 
    \end{tabular}
\end{table}

To further investigate the effectiveness of the components proposed in our study, we performed a series of ablation experiments. These experiments primarily aimed to evaluate the impact of multilevel feature fusion and the role of both the Prompt Semantic Supervision (PSS) module and the SMA module in AI-Generated video quality assessment.

Our model, MSA-VQA, employs a multilevel feature fusion strategy that efficiently captures frame-level local details and complex semantic variations across videos. As shown in Table \ref{tab:Ablation1}, integrating features across three different scales allows MSA-VQA to achieve state-of-the-art performance. The results underscore the importance of multilevel feature fusion in enhancing the model's capacity to comprehend and process both the fine-grained details within video frames and the broader semantic transitions between videos.

\begin{table}[h!]
    \centering
    \caption{Ablation study on the usage of Prompt Semantic Supervision Module and Semantic Mutation-aware Module.} 
    \label{tab:Ablation2}
    \begin{tabular}{cc|cc|c}
    \toprule 
    PSS & SMA & SRCC$\uparrow$ & PLCC$\uparrow$ & Avg Score$\uparrow$ \\
    \midrule 
    $\checkmark$ & $\checkmark$ & \textbf{0.810} & \textbf{0.825} & \textbf{0.818} \\
    $\checkmark$ & $\times$ & 0.786 & 0.808 & 0.796 \\
    $\times$ & $\times$ & 0.725 & 0.756 & 0.740 \\
    \bottomrule 
    \end{tabular}
\end{table}

The introduction of the PSS module was a critical enhancement to our model. As shown in Table \ref{tab:Ablation2}, incorporating the PSS module resulted in significant improvements across all evaluation metrics. Specifically, the Avg score increased from 0.740 to 0.796, underscoring the importance of maintaining consistency between AI-Generated videos and their corresponding prompts. This consistency is essential for accurately assessing video quality, as it directly impacts the relevance and coherence of the video content with respect to the given prompts.

Furthermore, the inclusion of the SMA module was pivotal in detecting semantic mutations within videos. By analyzing semantic shifts between frames, the SMA module enabled a more fine-grained evaluation of AI-Generated videos. The improvement in PLCC from 0.808 to 0.825 (+0.017) following the addition of the SMA module highlights its effectiveness in enhancing the overall quality of AI-Generated video assessments.

In summary, the ablation studies clearly demonstrate that both multilevel feature fusion and the specialized PSS and SMA modules contribute significantly to the performance of our AI-Generated video quality assessment model. These results emphasize the importance of a comprehensive approach that addresses various aspects of video content, from local details to global feature and semantic coherence, in order to improve the performance of AI-Generated Video Quality Assessment.

\section{CONCLUSION}
\label{sec:conclus}
\setlength{\parskip}{0pt}
In this paper, we introduce MSA-VQA, a novel model designed for evaluating the quality of AI-Generated videos. Our model utilizes a multilevel framework to analyze videos at the frame, segment, and full video levels, capturing both fine-grained details and high-level semantic content. By integrating the PSS and SMA modules, MSA-VQA ensures precise alignment with textual prompts and effectively detects semantic mutation shifts, leading to a more comprehensive video quality assessment. Extensive experiments demonstrate that our method achieves SOTA performance, while ablation studies validate the individual contributions of each component and highlight the overall superiority of our model.

\section{ACKNOWLEDGMENTS}
This work is supported by National Natural Science Foundation of China under grants 62302184.

\bibliographystyle{IEEEtran}
\small\bibliography{main}


\end{document}